\documentclass[10pt, a4paper]{article}\pdfoutput=1
\usepackage{lrec}
\usepackage{multibib}
\newcites{languageresource}{Language Resources}
\usepackage{graphicx}
\usepackage{tabularx}
\usepackage{todonotes}
\usepackage{adjustbox}
\usepackage{soul}
\usepackage{titlesec}
\titleformat{\section}{\normalfont\large\bf\center}{\thesection.}{1em}{}
\titleformat{\subsection}{\normalfont\SmallTitleFont\bf\raggedright}{\thesubsection.}{1em}{}
\titleformat{\subsubsection}{\normalfont\normalsize\bf\raggedright}{\thesubsubsection.}{1em}{}
\renewcommand\thesection{\arabic{section}}
\renewcommand\thesubsection{\thesection.\arabic{subsection}}
\renewcommand\thesubsubsection{\thesubsection.\arabic{subsubsection}}
\usepackage{epstopdf}
\usepackage[utf8]{inputenc}
\usepackage{booktabs}
\usepackage{hyperref}
\usepackage{xstring}
\usepackage{caption}

\usepackage{color}

\usepackage{amsmath}
\usepackage{enumitem}
\usepackage[T2A,T1]{fontenc}
\usepackage[russian,english]{babel}
\babeltags{ru=russian}

\definecolor{myred}{RGB}{29, 80, 185}
\definecolor{myblue}{RGB}{231, 76, 60}
\definecolor{mygreen}{RGB}{27, 174, 60}
\definecolor{myorange}{RGB}{153,153,0}

\title{Word Sense Disambiguation for 158 Languages using Word Embeddings Only}

\name{\parbox{\textwidth}{\centering Varvara Logacheva$^1$, Denis Teslenko$^2$, Artem Shelmanov$^1$, Steffen Remus$^3$,\\Dmitry Ustalov$^{4\star}$\thanks{$^\star$ Currently at Yandex.}, Andrey Kutuzov$^5$, Ekaterina Artemova$^6$,\\Chris Biemann$^3$, Simone Paolo Ponzetto$^4$, Alexander Panchenko$^1$}\medskip}

\address{
$^1$Skolkovo Institute of Science and Technology, Moscow, Russia \\
\texttt{v.logacheva@skoltech.ru} \\
$^2$Ural Federal University, Yekaterinburg, Russia \\
$^3$Universit\"at Hamburg, Hamburg, Germany \\
$^4$Universit\"at Mannheim, Mannheim, Germany \\
$^5$University of Oslo, Oslo, Norway \\ 
$^6$Higher School of Economics, Moscow, Russia
}

\abstract{
Disambiguation of word senses in context is easy for humans, but is a major challenge for automatic approaches. Sophisticated supervised and knowledge-based models were developed to solve this task. However, (i) the inherent Zipfian distribution of supervised training instances for a given word and/or (ii) the  quality of linguistic knowledge representations motivate the development of completely unsupervised and knowledge-free approaches to word sense disambiguation (WSD). They are particularly useful for under-resourced languages which do not have any resources for building either supervised and/or knowledge-based models. In this paper, we present a method that takes as input a standard pre-trained word embedding model and induces a fully-fledged word sense inventory, which can be used for disambiguation in context. We use this method to induce a collection of sense inventories for 158 languages on the basis of the original pre-trained fastText word embeddings by~\newcite{Grave:18}, enabling WSD in these languages. Models and system are available online.\\ \newline
\Keywords{word sense induction, word sense disambiguation, word embeddings, sense embeddings, graph clustering}}

\begin{document}

\maketitleabstract

\section{Introduction}

There are many polysemous words in virtually any language. If not treated as such, they can hamper the performance of all semantic NLP tasks \cite{Resnik:06}. Therefore, the task of resolving the polysemy and choosing the most appropriate meaning of a word in context has been an important NLP task for a long time. It is usually referred to as Word Sense Disambiguation (WSD) and aims at assigning meaning to a word in context. 

The majority of approaches to WSD are based on the use of knowledge bases, taxonomies, and other external manually built resources \cite{Moro:14,Upadhyay:18}. However, different senses of a polysemous word occur in very diverse contexts and can potentially be discriminated with their help. The fact that semantically related words occur in similar contexts, and diverse words do not share common contexts, is known as distributional hypothesis and underlies the technique of constructing word embeddings from unlabelled texts. 
The same intuition can be used to discriminate between different senses of individual words. There exist methods of training word embeddings that can detect polysemous words and assign them different vectors depending on their contexts \cite{Athiwaratkun:18,Jain:19}. Unfortunately, many wide-spread word embedding models, such as GloVe \cite{Pennington:14}, word2vec \cite{Mikolov:13}, fastText \cite{Bojanowski:17}, do not handle polysemous words. Words in these models are represented with single vectors, which were constructed from diverse sets of contexts corresponding to different senses. In such cases, their disambiguation needs knowledge-rich approaches.

We tackle this problem by suggesting a method of post-hoc unsupervised WSD. It does not require any external knowledge and can separate different senses of a polysemous word using only the information encoded in pre-trained word embeddings. We construct a semantic similarity graph for words and partition it into densely connected subgraphs. This partition allows for separating different senses of polysemous words. Thus, the only language resource we need is a large unlabelled text corpus used to train embeddings. This makes our method applicable to under-resourced languages. Moreover, while other methods of unsupervised WSD need to train embeddings from scratch, we perform retrofitting of sense vectors based on existing word embeddings.

We create a massively multilingual application for on-the-fly word sense disambiguation. When receiving a text, the system identifies its language and performs disambiguation of all the polysemous words in it based on pre-extracted word sense inventories. The system works for 158 languages, for which pre-trained fastText embeddings  available~\cite{Grave:18}.\footnote{The full list languages is available at \href{https://fasttext.cc}{fasttext.cc} and includes English and 157 other languages for which embeddings were trained on a combination of Wikipedia and CommonCrawl texts.} The created inventories are based on these embeddings. To the best of our knowledge, our system is the only WSD system for the majority of the presented languages. Although it does not match the state of the art for resource-rich languages, it is fully unsupervised and can be used for virtually any language.

The contributions of our work are the following:

\begin{itemize}[noitemsep]
    \item We release word sense inventories associated with fastText embeddings for 158 languages.    
    \item We release a system that allows on-the-fly word sense disambiguation for 158 languages.
    \item We present \texttt{egvi} (Ego-Graph Vector Induction), a new algorithm of unsupervised word sense induction, which creates sense inventories based on pre-trained word vectors.
\end{itemize}

\section{Related Work}

There are two main scenarios for WSD: the supervised approach that leverages training corpora explicitly labelled for word sense, and the knowledge-based approach that derives sense representation from lexical resources, such as WordNet \cite{Miller:95}. In the supervised case WSD can be treated as a classification problem. Knowledge-based approaches construct sense embeddings, i.e. embeddings that separate various word senses.  

SupWSD \cite{Papandrea:17} is a state-of-the-art system for \textbf{supervised WSD}. %, together with preprocessing and feature extraction.  
It makes use of linear classifiers and a number of features such as POS tags, surrounding words, local collocations, word embeddings, and syntactic relations. %, and allows to choose between two sense inventories, WordNet and BabelNet. 
GlossBERT model \cite{Huang:19}, which is another implementation of supervised WSD, achieves a significant improvement by leveraging gloss information. This model benefits from sentence-pair classification approach, introduced by \newcite{Devlin:19} in their BERT contextualized embedding model. The input to the model consists of a context (a sentence which contains an ambiguous word) and a gloss (sense definition) from WordNet.  
The context-gloss pair is concatenated through a special token (\texttt{[SEP]}) and classified as positive or negative.

On the other hand, \textbf{sense embeddings} are an alternative to traditional word vector models such as word2vec, fastText or GloVe, which represent %known to be 
 monosemous words well but fail for ambiguous words. Sense embeddings represent individual senses of polysemous words as separate vectors. They can be linked to an explicit inventory \cite{Iacobacci:15} or induce a sense inventory from unlabelled data \cite{Iacobacci:19}. 
LSTMEmbed \cite{Iacobacci:19} aims at learning sense embeddings linked to BabelNet \cite{Navigli:12:babelnet}, %sense inventory, 
at the same time handling word ordering, and using pre-trained embeddings as an objective. Although it was tested only on English, the approach can be easily adapted to other languages present in BabelNet. 
However, manually labelled datasets as well as knowledge bases exist only for a small number of well-resourced languages. Thus, to disambiguate polysemous words in other languages one has to resort to fully unsupervised techniques. 

The task of \textbf{Word Sense Induction} (WSI) can be seen as an unsupervised version of WSD. WSI aims at clustering word senses and does not require to map each cluster to a predefined sense. Instead of that, word sense inventories are induced automatically from the clusters,  treating each cluster as a single sense of a word. WSI approaches fall into three main groups: context clustering, word ego-network clustering and synonyms (or substitute) clustering. 

{\bf Context clustering} approaches consist in creating %are dominated by a series of increasingly sophisticated graphical models. For each word they create 
vectors which characterise words' contexts and clustering these vectors. 
Here, the definition of context may vary from window-based context to latent topic-alike context. Afterwards, the resulting clusters are either used as senses directly \cite{kutuzov2018russian}, or employed further to learn sense embeddings via Chinese Restaurant Process algorithm \cite{Li:15}, AdaGram, a Bayesian extension of the Skip-Gram model \cite{Bartunov:16}, AutoSense, an extension of the LDA topic model \cite{Amplayo:19}, and other techniques.  

{\bf Word ego-network clustering} is applied to semantic graphs. The nodes of a semantic graph are words, and edges between them denote semantic relatedness which is usually evaluated with cosine similarity of the corresponding embeddings \cite{Pelevina:16} or by PMI-like measures \cite{Hope:13:uos}. Word senses are induced via graph clustering algorithms, such as Chinese Whispers \cite{Biemann:06} or MaxMax \cite{Hope:13:maxmax}.
The technique suggested in our work belongs to this class of methods and is an extension of the method presented by \newcite{Pelevina:16}.

{\bf Synonyms and substitute clustering} approaches %benefit from various techniques which allow to obtain word synonyms or word substitutes. For each word, they 
create vectors which represent synonyms or substitutes of polysemous words. Such vectors are created using synonymy dictionaries \cite{Ustalov:19} or context-dependent substitutes obtained from a language model \cite{Amrami:18}. Analogously to previously described techniques, word senses are induced by clustering these vectors.

\section{Algorithm for Word Sense Induction}

The majority of word vector models do not discriminate between multiple senses of individual words. However, a polysemous word can be identified via manual analysis of its nearest neighbours---they reflect different senses of the word. Table \ref{tab:polysemous_contexts} shows manually sense-labelled most similar terms to the word \textit{Ruby} according to the pre-trained fastText model~\cite{Grave:18}. As it was  suggested early by \newcite{Widdows:02}, the distributional properties of a word can be used to construct a graph of words that are semantically related to it, and if a word is polysemous, such graph can easily be partitioned into a number of densely connected subgraphs corresponding to different senses of this word. Our algorithm is based on the same principle. 

\begin{table}
\begin{tabular}{p{8.1cm}}
\toprule

.Ruby, {\color{myred} CRuby}, {\color{myred} CoffeeScript}, {\color{myblue} Ember}, {\color{myorange} Faye}, {\color{myblue} Garnet}, {\color{myblue}  Gem}, {\color{myred} Groovy}, {\color{myred} Haskell}, {\color{mygreen} Hazel}, {\color{myred}  JRuby}, {\color{myorange} Jade}, {\color{myorange} Jasmine}, {\color{myorange} Josie}, {\color{myred} Jruby}, {\color{myorange} Lottie}, {\color{myorange} Millie}, {\color{myred} Oniguruma}, {\color{myblue} Opal}, {\color{myred} Python}, RUBY, Ruby., Ruby-like, 
{\color{myorange} Rabbitfoot}, {\color{myred} RubyMotion}, {\color{myred} Rails}, {\color{myred} Rubinius}, Ruby-, {\color{myred} Ruby-based}, {\color{myred} Ruby2}, {\color{myred} RubyGem}, {\color{myred} RubyGems}, {\color{myred} RubyInstaller}, {\color{myred}RubyOnRails}, {\color{myorange} RubyRuby}, {\color{myred} RubySpec}, {\color{myred} Rubygems}, {\color{myred} Rubyist}, {\color{myred} Rubyists}, Rubys, {\color{myorange} Sadie}, {\color{myblue} Sapphire}, {\color{myorange} Sypro}, {\color{mygreen} Violet}, {\color{myred} jRuby}, ruby, {\color{myred} rubyists}\\

\bottomrule
\end{tabular}
\caption{Top nearest neighbours of the fastText vector of the word \textit{Ruby} are clustered according to various senses of this word: {\color{myred} programming language}, {\color{myblue} gem}, {\color{myorange} first name}, {\color{mygreen} color}, but also its spelling variations (typeset in black color).}
\label{tab:polysemous_contexts}
\end{table}

\subsection{SenseGram: A Baseline Graph-based Word Sense Induction Algorithm}

SenseGram is the method proposed by \newcite{Pelevina:16} that separates nearest neighbours to induce word senses and constructs sense embeddings for each sense. It starts by constructing an \textit{ego-graph} (semantic graph centred at a particular word) of the word and its nearest neighbours. The edges between the words denote their semantic relatedness, e.g. the two nodes are joined with an edge if cosine similarity of the corresponding embeddings is higher than a pre-defined threshold. The resulting graph can be clustered into subgraphs which correspond to senses of the word. 

The sense vectors are then constructed by averaging embeddings of words in each resulting cluster. In order to use these sense vectors for word sense disambiguation in text, the authors compute the probabilities of sense vectors of a word given its context or the similarity of the sense vectors to the context.

\subsection{\texttt{egvi} (Ego-Graph Vector Induction): A Novel Word Sense Induction Algorithm}
\label{sec:egvi}

\paragraph{Induction of Sense Inventories}

One of the downsides of the described above algorithm is noise in the generated graph, namely, unrelated words and wrong connections. They hamper the separation of the graph. Another weak point is the imbalance in the nearest neighbour list, when a large part of it is attributed to the most frequent sense, not sufficiently representing the other senses. This can lead to construction of incorrect sense vectors.

We suggest a more advanced procedure of graph construction that uses the interpretability of vector addition and subtraction operations in word embedding space \cite{Mikolov:13} while the previous algorithm only relies on the list of nearest neighbours in word embedding space. The key innovation of our algorithm is the use of vector subtraction to find pairs of most dissimilar graph nodes and construct the graph only from the nodes included in such ``anti-edges''. Thus, our algorithm is based on \textit{graph-based} word sense induction, but it also relies on \textit{vector-based} operations between word embeddings to perform filtering of graph nodes. Analogously to the work of \newcite{Pelevina:16}, we construct a semantic relatedness graph from a list of nearest neighbours, but we filter this list using the following procedure:

\begin{enumerate}

\item Extract a list $\mathcal{N}$ = \{$w_{1}$, $w_{2}$, ..., $w_{N}$\} of $N$ nearest neighbours for the target (ego) word vector $w$.

\item Compute a list $\Delta$ = \{$\delta_{1}$, $\delta_{2}$, ..., $\delta_{N}$\} for each $w_{i}$ in $\mathcal{N}$, where $\delta_{i}~=~w-w_{i}$. The vectors in $\delta$ contain the components of sense of $w$ which are not related to the corresponding nearest neighbours from $\mathcal{N}$.

\item Compute a list $\overline{\mathcal{N}}$ = \{$\overline{w_{1}}$, $\overline{w_{2}}$, ..., $\overline{w_{N}}$\}, such that $\overline{w_{i}}$ is in the top nearest neighbours of $\delta_{i}$ in the embedding space. In other words, $\overline{w_{i}}$ is a word which is the most similar to the target (ego) word $w$ and least similar to its neighbour $w_{i}$. We refer to $\overline{w_{i}}$ as an \textit{anti-pair} of $w_{i}$. The set of $N$ nearest neighbours and their anti-pairs form a set of \textit{anti-edges} i.e. pairs of most dissimilar nodes -- those which should not be connected: $\overline{E} = \{ (w_{1},\overline{w_{1}}), (w_{2},\overline{w_{2}}), ..., (w_{N},\overline{w_{N}})\}$. 

To clarify this, consider the target (ego) word $w = \textit{python}$, its top similar term $w_1 = \textit{Java}$ and the resulting anti-pair $\overline{w_i} = \textit{snake}$ which is the top related term of $\delta_1 = w - w_1$. Together they form an anti-edge $(w_i,\overline{w_i})=(\textit{Java}, \textit{snake})$ composed of a pair of semantically dissimilar terms.

\item Construct $V$, the set of vertices of our semantic graph $G=(V,E)$ from the list of anti-edges $\overline{E}$, with the following recurrent procedure: $V = V \cup \{ w_{i}, \overline{w_{i}}: w_{i} \in \mathcal{N}, \overline{w_{i}} \in \mathcal{N}\}$, i.e. we add a word from the list of nearest neighbours \textit{and} its anti-pair only if both of them are nearest neighbours of the original word $w$. We do not add $w$'s nearest neighbours if their anti-pairs do not belong to $\mathcal{N}$. Thus, we add only words which can help discriminating between different senses of $w$. 

\item Construct the set of edges $E$ as follows. For each $w_{i}~\in~\mathcal{N}$ we extract a set of its $K$ nearest neighbours $\mathcal{N}'_{i} = \{u_{1}, u_{2}, ..., u_{K}\}$ and define $E = \{(w_{i}, u_{j}): w_{i}~\in~V, u_j~\in~V, u_{j}~\in~\mathcal{N}'_{i}, u_{j}~\neq~\overline{w_{i}}\}$. In other words, we remove edges between a word $w_{i}$ and its nearest neighbour $u_j$ if $u_j$ is also its anti-pair. According to our hypothesis, $w_{i}$ and $\overline{w_{i}}$ belong to different senses of $w$, so they should not be connected (i.e. we never add anti-edges into $E$). Therefore, we consider any connection between them as noise and remove it. 

\end{enumerate}

Note that $N$ (the number of nearest neighbours for the target word $w$) and $K$ (the number of nearest neighbours of $w_{ci}$) do not have to match. The difference between these parameters is the following. $N$ defines how many words will be considered for the construction of ego-graph. On the other hand, $K$ defines the degree of relatedness between words in the ego-graph --- if $K = 50$, then we will connect vertices $w$ and $u$ with an edge only if $u$ is in the list of 50 nearest neighbours of $w$. Increasing $K$ increases the graph connectivity and leads to lower granularity of senses.

According to our hypothesis, nearest neighbours of $w$ are grouped into clusters in the vector space, and each of the clusters corresponds to a sense of $w$. The described vertices selection procedure allows picking the most representative members of these clusters which are better at discriminating between the clusters. In addition to that, it helps dealing with the cases when one of the clusters is over-represented in the nearest neighbour list. In this case, many elements of such a cluster are not added to $V$ because their anti-pairs fall outside the nearest neighbour list. This also improves the quality of clustering.

After the graph construction, the clustering is performed using the Chinese Whispers algorithm \cite{Biemann:06}. This is a bottom-up clustering procedure that does not require to pre-define the number of clusters, so it can correctly process polysemous words with varying numbers of senses as well as unambiguous words.

\begin{figure}[ht]
	\centering
	\includegraphics[width=1.0\linewidth]{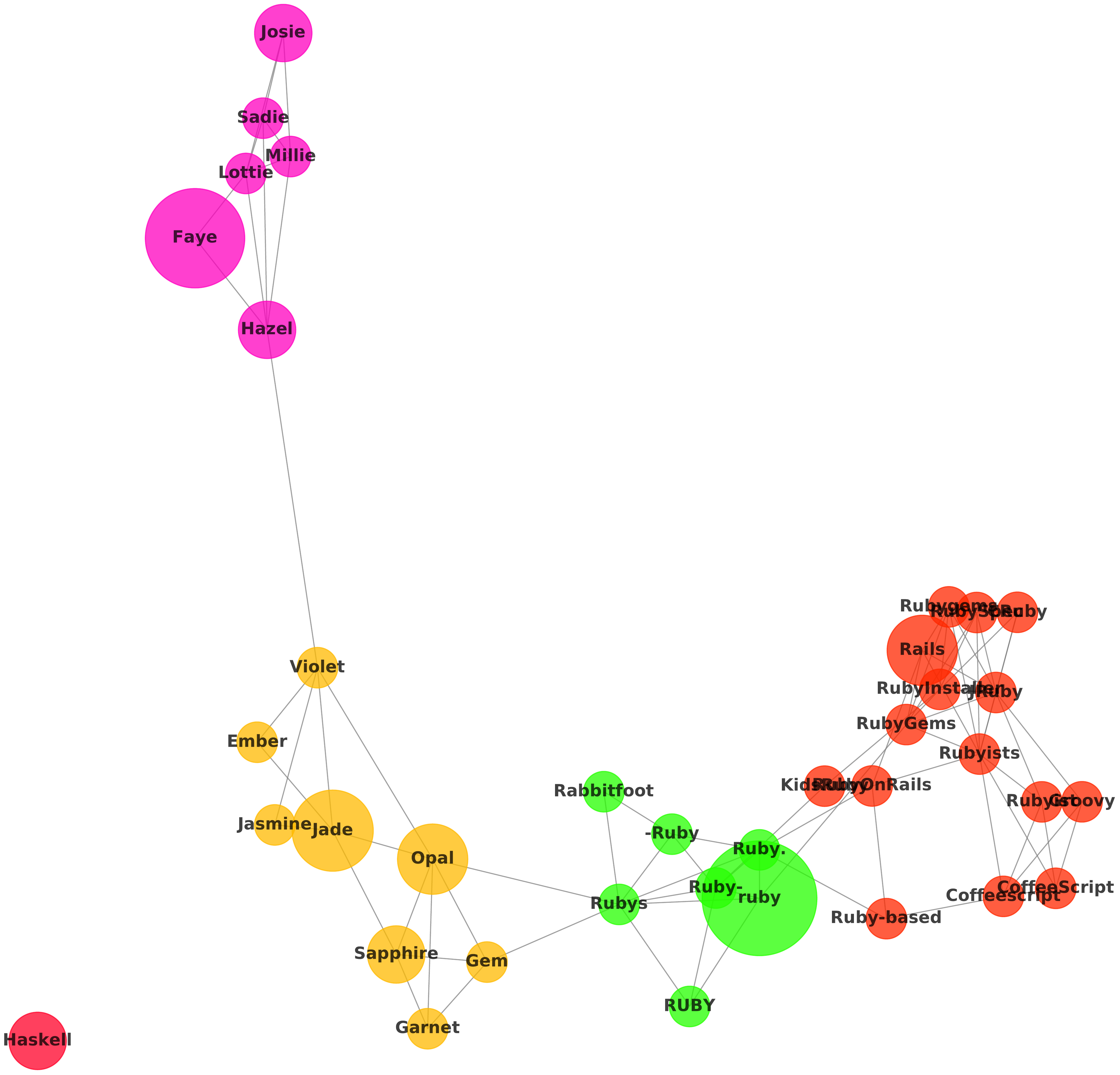} %crawl-300d-2M.vec.top50.inventory.tsv.Ruby-upper
	\caption{The graph of nearest neighbours of the word \textit{Ruby} can be separated according several senses: programming languages, female names, gems, as well as a cluster of different spellings of the word \textit{Ruby}. }
	\label{fig:table_ego_graph}
\end{figure}

Figure \ref{fig:table_ego_graph} shows an example of the resulting pruned graph of for the word \textit{Ruby} for $N = 50$ nearest neighbours in terms of the fastText cosine similarity. In contrast to the baseline method by~\cite{Pelevina:16} where all $50$ terms are clustered, in the method presented in this section we sparsify the graph by removing $13$ nodes which were not in the set of the ``anti-edges'' i.e.  pairs of most dissimilar terms out of these 50 neighbours. Examples of anti-edges i.e. pairs of most dissimilar terms for this graph include: (\textit{Haskell}, \textit{Sapphire}), (\textit{Garnet}, \textit{Rails}), (\textit{Opal}, \textit{Rubyist}), (\textit{Hazel}, \textit{RubyOnRails}), and (\textit{Coffeescript}, \textit{Opal}).

\paragraph{Labelling of Induced Senses}

We label each word cluster representing a sense to make them and the WSD results interpretable by humans. Prior systems used hypernyms to label the clusters~\cite{ruppert-etal-2015-jobimviz,panchenko-etal-2017-unsupervised}, e.g. ``animal'' in the ``python (animal)''. However, neither hypernyms nor rules for their automatic extraction are available for all 158 languages. Therefore, we use a simpler method to select a keyword which would help to interpret each cluster. For each graph node $v \in V$ we count the number of anti-edges it belongs to: $count(v) = | \{(w_i,\overline{w_i}) : (w_i,\overline{w_i}) \in \overline{E} \wedge (v = w_i \vee v = \overline{w_i})  \} |$.  A graph clustering yields a partition of $V$ into $n$ clusters: $V~=~\{V_1, V_2, ..., V_n\}$. For each cluster  $V_i$  we define a \textit{keyword} $w^{key}_i$ as the word with the largest number of anti-edges $count(\cdot)$ among words in this cluster.

\paragraph{Word Sense Disambiguation}
We use keywords defined above to obtain vector representations of senses. In particular, we simply use word embedding of the keyword $w^{key}_i$ as a sense representation $\mathbf{s}_i$ of the target word $w$ to avoid explicit computation of sense embeddings like in~\cite{Pelevina:16}. Given a sentence $\{w_1, w_2, ..., w_{j}, w, w_{j+1}, ..., w_n\}$ represented as a matrix of word vectors, we define the context of the target word $w$ as $\textbf{c}_w = \dfrac{\sum_{j=1}^{n} w_j}{n}$. Then, we define the most appropriate sense $\hat{s}$ as the sense with the highest cosine similarity to the embedding of the word's context:
$$\hat{s} = \underset{s_i}{\arg\max} \dfrac{\textbf{c}_w \cdot \mathbf{s}_i}{||\textbf{c}_w|| \cdot ||\mathbf{s}_i||}.$$

\section{System Design}

We release a system for on-the-fly WSD for 158 languages. Given textual input, it identifies polysemous words and retrieves senses that are the most appropriate in the context.

\begin{figure*}[ht!]
	\centering
    \includegraphics[width=0.7\linewidth]{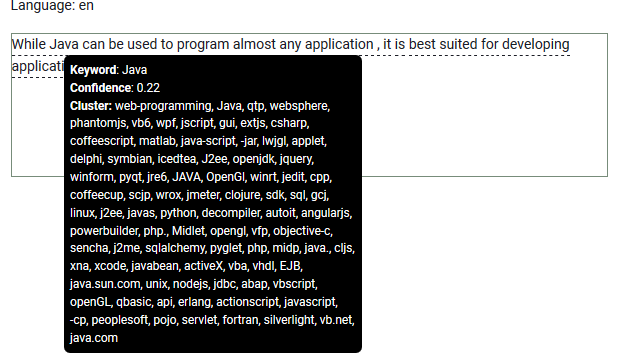}
    \caption{Interface of our WSD module with examples for the English language. Given a sentence, it identifies polysemous words and retrieves the most appropriate sense (labelled by the centroid word of a corresponding cluster).}
    \label{fig:interface}
\end{figure*}

\subsection{Construction of Sense Inventories} 

To build word sense inventories (sense vectors) for 158 languages, we utilised GPU-accelerated routines for search of similar vectors implemented in Faiss library \cite{Johnson:19}. The search of nearest neighbours takes substantial time, therefore, acceleration with GPUs helps to significantly reduce the word sense construction time. To further speed up the process, we keep all intermediate results in memory, which results in substantial RAM consumption of up to 200 Gb.

The construction of word senses for all of the 158 languages takes a lot of computational resources and imposes high requirements to the hardware. For calculations, we use in parallel 10--20 nodes of the Zhores cluster \cite{Zacharov:19} empowered with Nvidia Tesla V100 graphic cards. For each of the languages, we construct inventories based on 50, 100, and 200 neighbours for 100,000 most frequent words. The vocabulary was limited in order to make the computation time feasible. The construction of inventories for one language takes up to $10$ hours, with $6.5$ hours on average. Building the inventories for all languages took more than 1,000 hours of GPU-accelerated computations.% (for a single node).

We release the constructed sense inventories for all the available languages. They contain all the necessary information for using them in the proposed WSD system or in other downstream tasks.

\subsection{Word Sense Disambiguation System}

The first text pre-processing step is language identification, for which we use the fastText language identification models by \newcite{Bojanowski:17}. Then the input is tokenised. 
For languages which use Latin, Cyrillic, Hebrew, or Greek scripts, we employ the Europarl tokeniser.\footnote{\url{https://www.statmt.org/europarl}} For Chinese, we use the Stanford Word Segmenter \cite{Tseng:05}. For Japanese, we use Mecab \cite{Kudo:06}. We tokenise Vietnamese with UETsegmenter \cite{Nguyen:16}. All other languages %\footnote{The full list is available at \url{https://github.com/uhh-lt/158/blob/master/158_tokenizer/icu_langs.tsv}.} 
are processed with the ICU tokeniser, as implemented in the PyICU project.\footnote{\url{https://pypi.org/project/PyICU}}
After the tokenisation, the system analyses all the input words with pre-extracted sense inventories and defines the most appropriate sense for polysemous words.

Figure \ref{fig:interface} shows the interface of the system. It has a textual input form. The automatically identified language of text is shown above. A click on any of the words displays a prompt (shown in black) with the most appropriate sense of a word in the specified context and the confidence score. In the given example, the word \textit{Jaguar} is correctly identified as a car brand. This system is based on the system by~\newcite{Ustalov:18}, extending it with a back-end for multiple languages, language detection, and sense browsing capabilities.

\section{Evaluation}

We first evaluate our converted embedding models on multi-language lexical similarity and relatedness tasks, as a sanity check, to make sure the word sense induction process did not hurt the general performance of the embeddings. Then, we test the sense embeddings on WSD task.

\subsection{Lexical Similarity and Relatedness}
\paragraph{Experimental Setup}

We use the SemR-11 datasets\footnote{\small\url{https://github.com/Lambda-3/Gold-Standards/tree/master/SemR-11}} \cite{Barzegar:18}, which contain word pairs with manually assigned similarity scores from 0 (words are not related) to 10 (words are fully interchangeable) for 12 languages: English (en), Arabic (ar), German (de), Spanish (es), Farsi (fa), French (fr), Italian (it), Dutch (nl), Portuguese (pt), Russian (ru), Swedish (sv), Chinese (zh). 
The task is to assign relatedness scores to these pairs so that the ranking of the pairs by this score is close to the ranking defined by the oracle score. The performance is measured with Pearson correlation of the rankings.
Since one word can have several different senses in our setup, we follow \newcite{Remus:18} and define the relatedness score for a pair of words as the \textbf{maximum cosine similarity} between any of their sense vectors.  

We extract the sense inventories from fastText %\footnote{\url{https://fasttext.cc}} 
embedding vectors. 
We set $N=K$ %(the number of nearest neighbours for the target word) to 200 
for all our experiments, i.e. the number of vertices in the graph and the maximum number of vertices' nearest neighbours match. %We vary the parameter $K$ (the number of nearest neighbours of the vertices of the ego-graph): 
We conduct experiments with $N=K$ set to 50, 100, and 200. 
For each cluster $V_i$ we create a sense vector $s_i$ by averaging vectors that belong to this cluster. We rely on the methodology of~\cite{Remus:18} shifting the generated sense vector to the direction of the original word vector: $s_i~=~\lambda~w + (1-\lambda)~\dfrac{1}{n}~\sum_{u~\in~V_i} cos(w, u)\cdot u, $ where, $\lambda \in [0, 1]$, $w$ is the embedding of the original word, $cos(w, u)$ is the cosine similarity between $w$ and $u$, and $n=|V_i|$. %The intuition behind this approach is that since clusters $V_i$ do not contain the word $w$ itself, their centroids can be situated too far from it.
By introducing the linear combination of $w$ and $u~\in~V_i$ we enforce the similarity of sense vectors to the original word important for this task. In addition to that, we weight $u$ by their similarity to the original word, so that more similar neighbours contribute more to the sense vector. The shifting parameter $\lambda$ is set to $0.5$, following \newcite{Remus:18}.

A fastText model is able to generate a vector for each word even if it is not represented in the vocabulary, due to the use of subword information. However, our system cannot assemble sense vectors for out-of-vocabulary words, for such words it returns their original fastText vector. Still, the coverage of the benchmark datasets by our vocabulary is at least 85\% and approaches 100\% for some languages, so we do not have to resort to this back-off strategy very often.

We use the original fastText vectors as a \textbf{baseline}. In this case, we compute the relatedness scores of the two words as a cosine similarity of their vectors.

\paragraph{Discussion of Results}

\begin{figure}[!ht]
	\centering
    \includegraphics[width=1.0\linewidth]{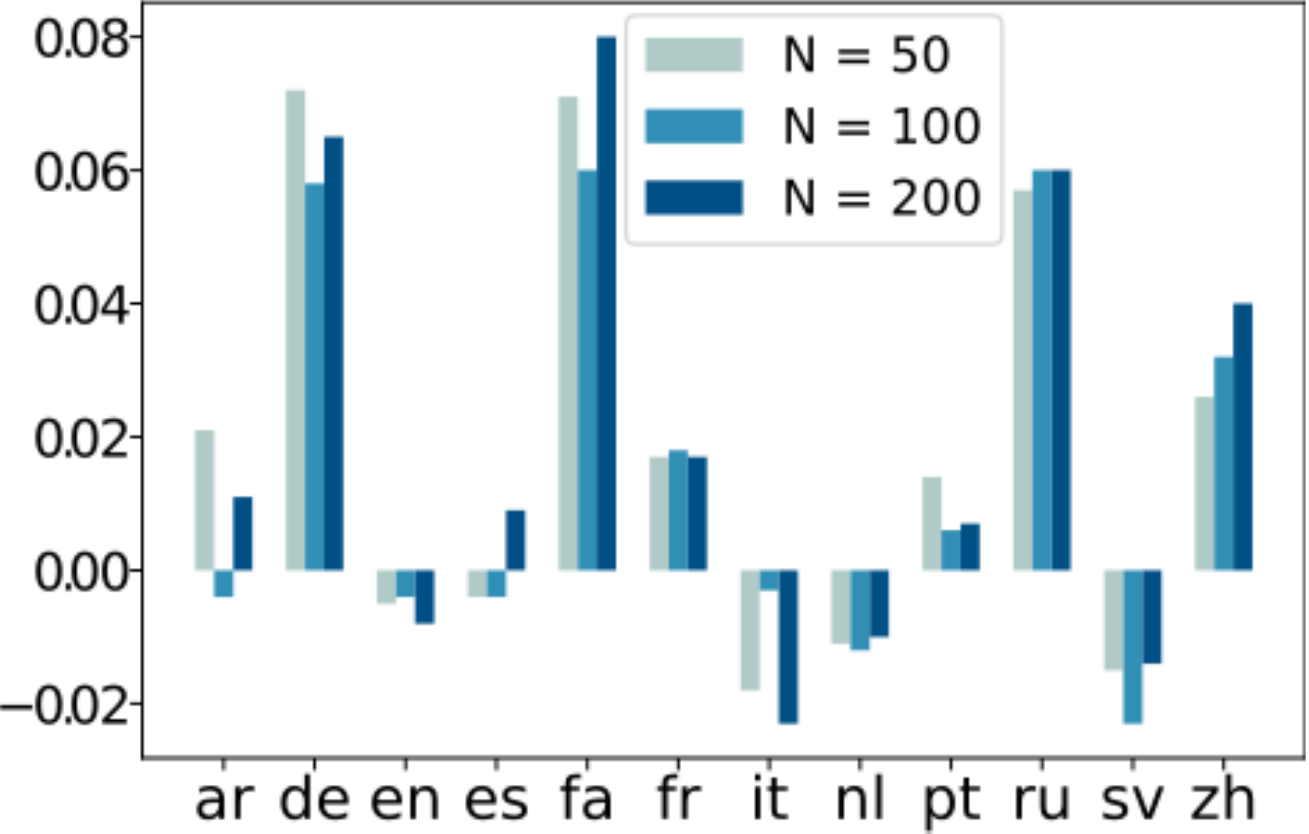}
    \caption{Absolute improvement of Pearson correlation scores of our embeddings compared to fastText. This is the averaged difference of the scores for all word similarity benchmarks.}
    \label{fig:all_lang_result}
\end{figure}

We compute the relatedness scores for all benchmark datasets using our sense vectors and compare them to cosine similarity scores of original fastText vectors. The results vary for different languages. Figure \ref{fig:all_lang_result} shows the change in Pearson correlation score when switching from the baseline fastText embeddings to our sense vectors. % (the scores are averaged for all SemR-11 datasets). 
The new vectors significantly improve the relatedness detection for German, Farsi, Russian, and Chinese, whereas for Italian, Dutch, and Swedish the score slightly falls behind the baseline. For other languages, the performance of sense vectors is on par with regular fastText. %We also notice that $K = 50$ often returns better result, and for Italian and Dutch leads to smaller drop of quality.

\subsection{Word Sense Disambiguation}
The purpose of our sense vectors is disambiguation of polysemous words. Therefore, we test the inventories constructed with \texttt{egvi} on the Task 13 of SemEval-2013 --- Word Sense Induction \cite{jurgens-klapaftis-2013-semeval}. The task is to identify the different senses of a target word in context in a fully unsupervised manner. 

\paragraph{Experimental Setup}

The dataset %of the task 
consists of a set of polysemous words: 20 nouns, 20 verbs, and 10 adjectives and specifies 20 to 100 contexts per word, with the total of 4,664 contexts, drawn from %. The contexts were drawn from 
the Open American National Corpus. Given a set of contexts of a polysemous word, the participants of the competition had to divide them into clusters by sense of the word. The contexts are manually labelled with WordNet senses of the target words, the gold standard clustering is generated from this labelling.

The task allows two setups: \textit{graded} WSI where participants can submit multiple senses per word and provide the probability of each sense in a particular context, and \textit{non-graded} WSI where a model determines a single sense for a word in context. In our experiments we performed \textit{non-graded} WSI. We considered the most suitable sense as the one with the highest cosine similarity with embeddings of the context, as described in Section \ref{sec:egvi}.

The performance of WSI models is measured with three metrics that require mapping of sense inventories (Jaccard Index, Kendall's $\tau$, and WNDCG) and two cluster comparison metrics (Fuzzy NMI and  Fuzzy B-Cubed).

\begin{table*}[ht!]
\footnotesize
\centering
\begin{tabular}{l|rrr|rr}
\toprule
  & \multicolumn{3}{c}{Supervised Evaluation} & \multicolumn{2}{c}{ Clustering Evaluation}  \\
 \bf Model & \bf Jacc. Ind. & \bf Kendall's $\tau$ & \bf WNDCG & \bf F.NMI & \bf F.B-Cubed \\
\midrule
\multicolumn{6}{c}{\textbf{Baselines}} \\
\midrule
One sense for all & 0.192 & 0.609 & 0.288 & 0.000 & 0.631 \\
One sense per instance & 0.000 & 0.000 & 0.000 & 0.071 & 0.000 \\
Most Frequent Sense & 0.455 & 0.465 & 0.339 &  -- & -- \\
\midrule
\multicolumn{6}{c}{\textbf{SemEval-2013 participants}}\\
\midrule
AI-KU (base) & 0.197 & 0.620 & 0.387 & 0.065 & 0.390 \\
%AI-KU (add1000) & 0.176 & 0.609 & 0.205 & 0.033 & 0.317 \\
AI-KU (remove5-add1000) & 0.244 & 0.642 & 0.332 & 0.039 & 0.451 \\
%Unimelb (5p) & 0.198 & 0.623 & 0.374 & 0.056 & 0.475 \\
Unimelb (50k) & 0.213 & 0.620 & 0.371 & 0.060 & 0.483 \\
%UoS (\#WN senses) & 0.171 & 0.600 & 0.298 & 0.046 & 0.186 \\
%UoS (top-3) & 0.220 & 0.637 & 0.370 & 0.044 & 0.451 \\
%La Sapienza (1) & 0.131 & 0.544 & 0.332 & --  & -- \\
%La Sapienza (2) & 0.131 & 0.535 & 0.394 & -- & -- \\
\midrule
\multicolumn{6}{c}{\textbf{Sense embeddings}}\\
\midrule
AdaGram, $\alpha$ = 0.05, 100 dim. vectors & 0.274 & 0.644  & 0.318  & 0.058  & 0.470  \\
%\midrule
%\multicolumn{6}{c}{\textbf{Sensegram} \cite{Pelevina:16}} \\
%\midrule
SenseGram (word2vec)   & 0.197 & 0.615 & 0.291 & 0.011 & 0.615 \\
%w2v   -- weighted -- sim. -- filter ($p=2$):   nouns & 0.179 & 0.626 & 0.304 & 0.011 & 0.623 \\
%JBT  -- weighted -- sim. -- filter ($p=2$)  & 0.205 & 0.624 & 0.291 & 0.017 & 0.598\\
%JBT  -- weighted -- sim. -- filter ($p=2$): nouns & 0.198 & 0.643 & 0.310 & 0.031 & 0.595\\
%TWSI -- weighted -- sim. -- filter ($p=2$): nouns & 0.215 & 0.651 & 0.318 & 0.030 & 0.573 \\
%\midrule
\texttt{\textbf{egvi}} (fastText, K=200) & 0.229 & 0.625 & 0.300 & 0.035 & 0.541 \\
\bottomrule
\end{tabular}
\caption{WSD performance on the SemEval-2013 Task 13 dataset for the English language.}
%  Model name denotes its configuration: word similarity graph -- pooling method -- disambiguation method -- context filtering. 
\label{tab:wsd}
\end{table*}

\paragraph{Discussion of Results}

We compare our model with the models that participated in the task, the baseline ego-graph clustering model by \newcite{Pelevina:16}, and AdaGram \cite{Bartunov:16}, a method that learns sense embeddings based on a Bayesian extension of the Skip-gram model. 
Besides that, we provide the scores of the simple \textbf{baselines} originally used in the task: assigning one sense to all words, assigning the most frequent sense to all words, and considering each context as expressing a different sense. The evaluation of our model was performed using the open source {\tt context-eval} tool.\footnote{\url{https://github.com/uhh-lt/context-eval}}

Table \ref{tab:wsd} shows the performance of these models on the SemEval dataset. Due to space constraints, we only report the scores of the best-performing SemEval participants, please refer to \newcite{jurgens-klapaftis-2013-semeval} for the full results. The performance of AdaGram and SenseGram models is reported according to \newcite{Pelevina:16}.

The table shows that the performance of \texttt{egvi} is similar to state-of-the-art word sense disambiguation and word sense induction models. In particular, we can see that it outperforms SenseGram on the majority of metrics. We should note that this comparison is not fully rigorous, because SenseGram induces sense inventories from word2vec as opposed to fastText vectors used in our work. %The results vary depending on the metrics, but  \texttt{egvi} is always in the top cohort.

\subsection{Analysis}

In order to see how the separation of word contexts that we perform corresponds to actual senses of polysemous words, we visualise ego-graphs produced by our method. Figure \ref{fig:table_ego_graph} shows the nearest neighbours clustering for the word \textit{Ruby}, which divides the graph into five senses:
\textit{Ruby-related programming tools}, e.g. RubyOnRails (orange cluster), \textit{female names}, e.g. Josie (magenta cluster), \textit{gems}, e.g. Sapphire (yellow cluster), \textit{programming languages in general}, e.g. Haskell (red cluster). Besides, this is typical for fastText embeddings featuring sub-string similarity, one can observe  a cluster of different spelling of the word \textit{Ruby} in green.

\begin{figure}[!ht]
	\centering
	\includegraphics[width=1.0\linewidth]{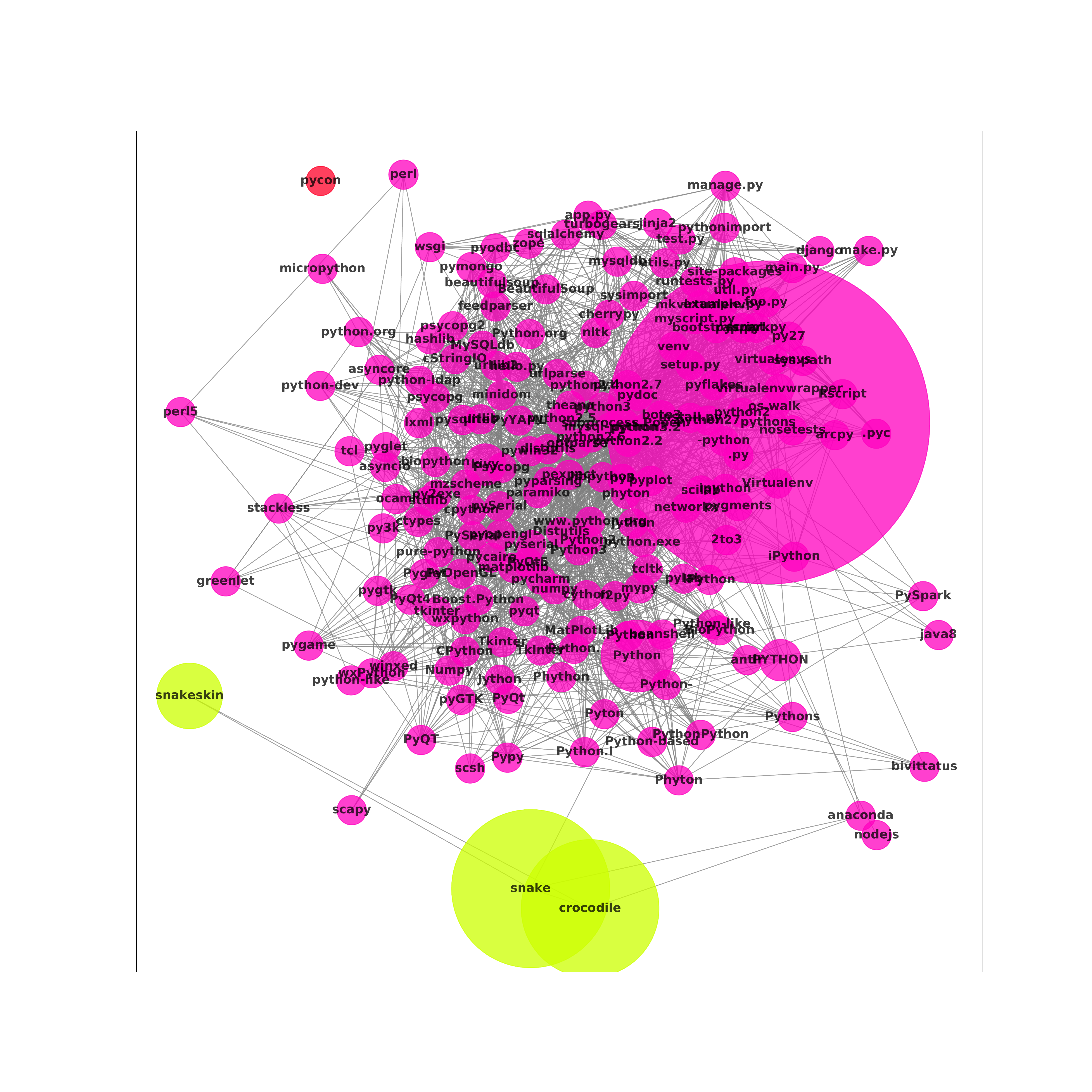}
	    \caption{Ego-graph for a polysemous word \textit{python} which is clustered into senses \textit{snake} (yellow),  \textit{programming language} (magenta), and \textit{conference} (red). Node size denotes word importance with the largest node in the cluster being used as a keyword to interpret an induced word sense. }
	    \label{fig:python_senses}
\end{figure}

Analogously, the word \textit{python} (see Figure \ref{fig:python_senses}) is divided into the senses of \textit{animals}, e.g. crocodile (yellow cluster), \textit{programming languages}, e.g. perl5 (magenta cluster), and \textit{conference}, e.g. pycon (red cluster). 

\begin{figure*}[!ht]
	\centering
	 \includegraphics[width=1.0\linewidth]{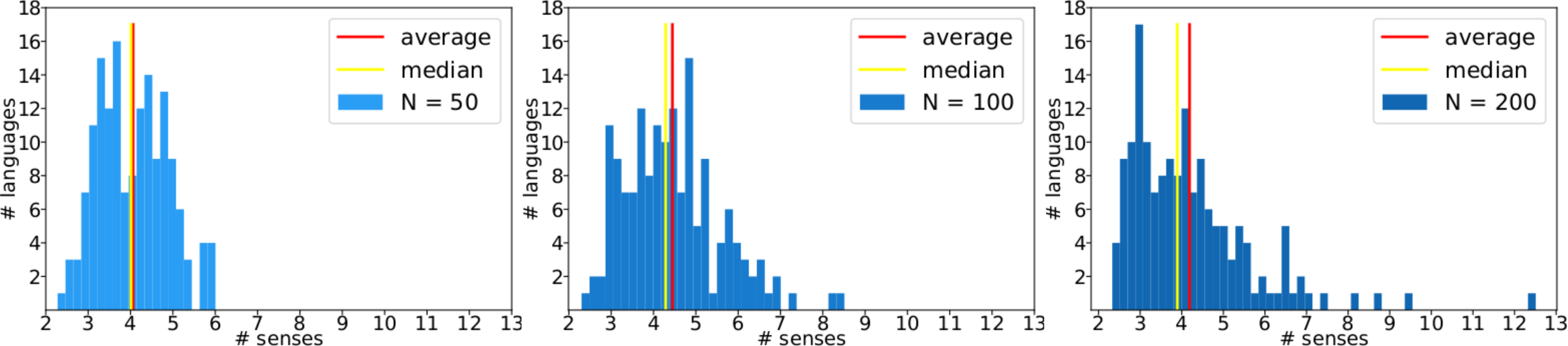}
	   \caption{Distribution of the number of senses per word in the generated inventories for all 158 languages for the number of neighbours set to: $N \in \{50, 100, 200\}$, $K \in \{50, 100, 200\}$ with $N = K$.}
	    \label{fig:n_senses}
\end{figure*}

In addition, we show a qualitative analysis of senses of \textit{mouse} and \textit{apple}. Table \ref{tab:senses_examples} shows nearest neighbours of the original words separated into clusters (labels for clusters were assigned manually). These inventories demonstrate clear separation of different senses, although it can be too fine-grained. For example, the first and the second cluster for \textit{mouse} both refer to computer mouse, but the first one addresses the different types of computer mice, and the second one is used in the context of mouse actions. Similarly, we see that \textit{iphone} and \textit{macbook} are separated into two clusters.
Interestingly, fastText handles typos, code-switching, and emojis by correctly associating all non-standard variants to the word they refer, and our method is able to cluster them appropriately. %Another strong point of our sense separation is that it is not inclined to merge different senses into one cluster.

Both inventories were produced with $K=200$, which ensures stronger connectivity of graph. However, we see that this setting still produces too many clusters. We computed the average numbers of clusters produced by our model with $K=200$ for words from the word relatedness datasets and compared these numbers with the number of senses in WordNet for English and RuWordNet \cite{Loukachevitch:14} for Russian (see Table \ref{tab:n_senses}). We can see that the number of senses extracted by our method is consistently higher than the real number of senses. 

We also compute the average number of senses per word for all the languages and different values of $K$ (see Figure \ref{fig:n_senses}). The average across languages does not change much as we increase $K$. %increases from $4.1$ to $4.5$ as we switch from $K=50$ to $K=100$ and then goes back to $4.2$ for $K=200$. 
However, for larger $K$ the average exceed the median value, %the distance between the average and the median grows, 
%does not change much as we increase $K$. However, as we expected, the distribution becomes more skewed towards smaller number of senses --- the median changes from $4.0$ for $K=50$ to $???$ for $K=200$ 
indicating that more languages have lower number of senses per word. 
At the same time, while at smaller $K$ the maximum average number of senses per word does not exceed $6$, larger values of $K$ produce outliers, e.g. English with $12.5$ senses.

Notably, there are no languages with an average number of senses less than $2$, while numbers on English and Russian WordNets are considerably lower. %polysemous words are less common than monosemous. 
This confirms that our method systematically over-generates senses. The presence of outliers shows that this effect cannot be eliminated by further increasing $K$, because the $i$-th nearest neighbour of a word for $i>200$ can be only remotely related to this word, even if the word is rare. Thus, our sense clustering algorithm needs a method of merging spurious senses.

\begin{table}[htbp]

\begin{adjustbox}{width=.5\textwidth,center}
\begin{tabular}{l|llll|l}
\toprule
          & mc  & rg  & simlex & ws353 & total                                                               \\ \midrule
          & \multicolumn{5}{c}{English}                                                                      \\ \midrule
inventory & 9.8 & 9.8 & 12.6   & 11.3  & 12.5                                                                \\ \midrule 
WordNet   & 3.6 & 3.7 & 6.5    & 5.5   & \begin{tabular}[c]{@{}l@{}}1.23 (nouns)\\ 2.16 (verbs)\end{tabular} \\ \midrule
          & \multicolumn{5}{c}{Russian}                                                                      \\ \midrule
inventory & 1.8 & 2.0 & --     & 2.2   & 2.97                                                                \\ \midrule
RuWordNet & 1.4 & 1.4 & --     & 1.8   & \begin{tabular}[c]{@{}l@{}}1.12 (nouns)\\ 1.33 (verbs)\end{tabular} \\ \bottomrule
\end{tabular}
\end{adjustbox}
\caption{Average number of senses for words from SemR-11 dataset in our inventory and in WordNet for English and ruWordNet for Russian. The rightmost column gives the average number of senses in the inventories and WordNets.}
\label{tab:n_senses}

\end{table}

\begin{table*}
\centering 
\small 

\begin{tabular}{p{2cm}|p{14.2cm}}
\toprule

\textbf{Label}   & \textbf{Nearest neighbours} \\ \bottomrule

\multicolumn{2}{c}{MOUSE} \\  \bottomrule

computer mouse types & touch-pad, logitec, scrollwheel, \textbf{mouses}, mouse.It, mouse.The, joystick, trackpads, nano-receiver, 800DPI, nony, track-pad, M325, keyboard-controlled, Intellipoint, MouseI, intellimouse, Swiftpoint, Evoluent, 800dpi, moused, game-pad, steelseries, ErgoMotion, IntelliMouse, \textless{}...\textgreater{} \\ \midrule

computer mouse actions & Ctrl-Alt, right-mouse, cursor, left-clicks, spacebar, rnUse, mouseclick, \textbf{click}, mousepointer,  keystroke, cusor, mousewheel, MouseMove, mousebutton, leftclick, click-dragging, mouse-button, cursor., arrow-key, double-clicks, mouse-down, ungrab, mouseX, arrow-keys, right-button, \textless{}...\textgreater{} \\ \midrule

rodent & Rodent, rodent, mousehole, rats, \textbf{mice}, mice-, hamster, SOD1G93A, meeses, mice.The, PDAPP, hedgehog, Maukie, rTg4510, mousey, meeces, rodents, cat, White-footed, rat, Mice, \textless{}...\textgreater{} \\ \midrule

keyboard & keyborad, keybard, keybord, keyboardThe, keyboad, keyboar, \textbf{Keyboard}, keboard, keyboardI, keyb, keyboard.This, keybaord, keyboard \\ \midrule

medical & SENCAR, mTERT, \textbf{mouse-specific} \\ \midrule

Latin & Apodemus, \textbf{Micormys} \\ \midrule

Latin & \textbf{Akodon} \\ \bottomrule 

\multicolumn{2}{c}{APPLE} \\ \bottomrule

iphone & mobileme, idevice, carplay, iphones, icloud, iwatch, ios5, ipod, \textbf{iphone}, android, ifans, iphone.I, iphone4, iphone5s, idevices, ipad, ios, ipad., iphone5, iphone., ios7 \\ \midrule

fruit & \textbf{apples}, apple-producing, Honeycrisp, apple-y, Macouns, apple-growing, pear, apple-pear, Gravensteins, apple-like, Apples, Honeycrisps, apple-related, Borkh, Braeburns, Starkrimson, Apples-, SweeTango, Elstar \\  \midrule

macbook & macbook, macbookpro, macbookair, imac, ibooks, \textbf{tuaw}, osx, macintosh, imacs, apple.com, applestore, Tagsapple, stevejobs, applecare \\ \midrule

fruit, typos &  blackerry, blackberry, blueberry, aplle, cidar, \textbf{apple.The}, apple.I, aple, appple, calvados, pie.It, \\ pinklady \\ \midrule

tokenisation issues, typos & Apple.This, AMApple, it.Apple, too.Apple, AppleApple, \textbf{up.Apple}, AppleA, Aplle, Apple.Apple \\ \midrule 

Apple criticism & anti-apple, Aple, Crapple, isheep, iDiots, crapple, Appple, iCrap, \textbf{non-apple} \\ \midrule

Bible & \textbf{Adam} \\ \midrule

cooking & \textbf{caramel-dipped} \\ \midrule

iphone & \textbf{earpod} \\ \midrule

Russian & \textbf{\textru{яблоко}} [Russian: ``apple''] \\ \midrule
 
emoji & \textbf{[apple emoji]} \\ \bottomrule

\end{tabular}
\caption{Clustering of senses for words \textit{mouse} and \textit{apple} produced by our method. Cluster labels in this table were assigned manually for illustrative purposes. For on-the-fly disambiguation we use centroid words in clusters as sense labels (shown here \textbf{in bold}).}
\label{tab:senses_examples}
\end{table*}

\section{Conclusions and Future Work}

We present \texttt{egvi}, a new algorithm for word sense induction based on graph clustering that is fully unsupervised and relies on graph operations between word vectors. We apply this algorithm to a large collection of pre-trained fastText word embeddings, releasing sense inventories for 158 languages.\footnote{Links to the produced datasets, online demo, and source codes are available at: \url{http://uhh-lt.github.io/158}.} These inventories contain all the necessary information for constructing sense vectors and using them in downstream tasks. The sense vectors for polysemous words can be directly retrofitted with the pre-trained word embeddings and do not need any external resources. As one application of these multilingual sense inventories, we present a multilingual word sense disambiguation system that performs unsupervised and knowledge-free WSD for 158 languages without the use of any dictionary or sense-labelled corpus.

The evaluation of quality of the produced sense inventories is performed on multilingual word similarity benchmarks, showing that our sense vectors improve the scores compared to non-disambiguated word embeddings. Therefore, our system in its present state can improve WSD and downstream tasks for languages where knowledge bases, taxonomies, and annotated corpora are not available and supervised WSD models cannot be trained.

A promising direction for future work is combining  distributional information from the induced sense inventories with lexical knowledge bases to improve WSD performance. Besides, we encourage the use of the produced word sense inventories in other downstream tasks. 

\section{Acknowledgements}

We acknowledge the support of the Deutsche Forschungsgemeinschaft (DFG) foundation under the ``JOIN-T 2'' and ``ACQuA'' projects. Ekaterina Artemova was supported by the framework of the HSE University Basic Research Program and Russian Academic  Excellence Project ``5-100''.

\section{Bibliographical References}

\bibliographystyle{lrec}
\bibliography{lrec2020-158}

\end{document}